# Improving Efficiency of DNN-based Relocalization Module for Autonomous Driving with Server-side Computing

Dengbo Li[1], Jieren Cheng[2,*], and Bernie Liu[3]

[1] School of Computer Science and Technology, Hainan University,
Haikou,570228, China
[2] Hainan blockchain technology engineering research center
Haikou,570228, China
[3] Hainan Shuyi Technology Co., Ltd.
Sanya,572025, China
`cjr22@163.com`

**Abstract.** In this work, we present a novel framework for camera relocation in autonomous vehicles, leveraging deep neural networks (DNN). While existing literature offers various DNN-based camera relocation methods, their deployment is hindered by their high computational demands during inference. In contrast, our approach addresses this challenge through edge cloud collaboration. Specifically, we strategically offload certain modules of the neural network to the server and evaluate the inference time of data frames under different network segmentation schemes to guide our offloading decisions. Our findings highlight the vital role of server-side offloading in DNN-based camera relocation for autonomous vehicles, and we also discuss the results of data fusion. Finally, we validate the effectiveness of our proposed framework through experimental evaluation.

## 1 Introduction

Autonomous driving technology has long been a challenging subject for both academia and the automotive industry. To ensure a safe, stable, and comfortable driving experience, autonomous vehicles require highly precise environmental mapping and accurate self-positioning. Nevertheless, environmental factors can interfere with the accuracy of laser radar, and GPS signals can be interrupted in certain scenarios, which limits their effectiveness in locating autonomous vehicles. As a result, visual simultaneous localization and mapping (SLAM) technology has emerged as a highly effective and viable solution. Through the use of visual sensors, SLAM technology enables the resolution of challenges surrounding robot positioning and map construction in unfamiliar environments. Relocalization is a crucial module in SLAM systems, enabling autonomous vehicles to obtain their precise pose under a high-precision map.

The traditional relocalization method involves building a feature descriptor database based on the existing map data and matching the query image with the feature descriptor [1,2,3]. Currently, many efficient retrieval methods exist, such as deep learning-based descriptor construction [4,5], word bag models [6,7], and VLAD [8]. Previous researchers have used priority correspondence search and 3D point division to improve positioning efficiency [9,10]. Additionally, they have utilized SIFT features and correspondence between 3D points and 2D points in the scene to enhance positioning accuracy [11,12].The traditional camera relocalization approach relies heavily on shallow feature information in the scene, which can lead to poor relocalization accuracy, large drift, and other failures.

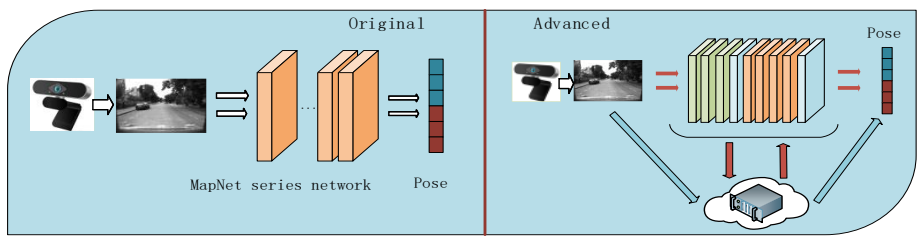

**Fig. 1.** Original: The relocation framework for the MapNet series. Feed the environmental photos collected by the sensors into the network for reasoning, and finally obtain the pose. Advanced: The upgraded version of the relocation framework splits the network, uploads the portion that consumes computing resources to the server for calculation, and finally returns the results to the mobile terminal for subsequent processing(The network types available for both frameworks are PoseNet-new version, MapNet, and MapNet++).

In recent years, there has been a growing interest in using deep learning-based scene construction models to perform pose regression. PoseNet [13] was the first to propose using the GoogleNet [14] depth neural network to directly return the 6-DoF pose of an input image. This method leverages the powerful feature extraction capabilities of neural networks to capture contextual features. The accuracy of pose estimation can be further improved by adding uncertainty measures and geometric loss to the PoseNet framework [15, 16].Laskar et al. [17] proposed using Siamese CNN to predict the relative position and pose of an input image, and then calculating the absolute position and pose based on the relative position and pose. This approach decouples the CNN and the scene coordinate system, greatly improving the model's generalization ability across different datasets. RobustLoc [18] obtained robustness to environmental disturbances from differential equations, extracted feature maps using CNN, and estimated vehicle attitude using a branch attitude decoder with multi-layer training. This method achieved robust performance in various environments. Simply using sparse descriptors can also regress scene coordinates without the need for dense RGB images [19]. The basic connection and coupling between multiple tasks can improve the model's understanding of the scene, further improving positioning performance [20, 21].

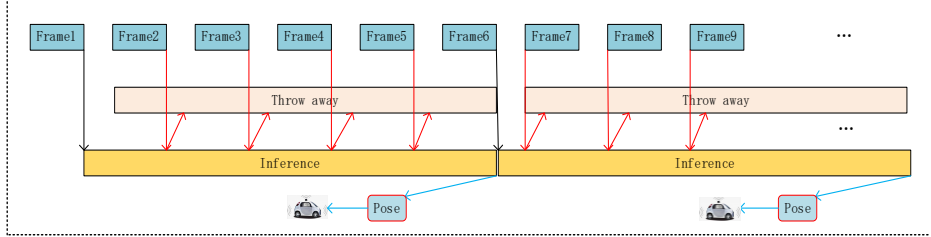

**Fig. 2.** The simulation pipeline of the original relocation framework, with red lines indicating discarded frames. Due to the huge amount of time spent in reasoning each frame, the environment frames captured by the sensor will be discarded during this period, and the vehicle will only receive a small amount of pose information.

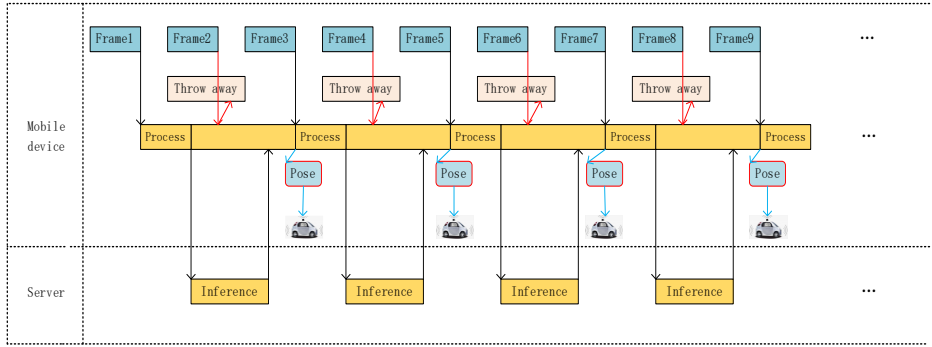

**Fig. 3.** The upgraded version repositions the simulation pipeline of the framework, with red lines indicating discarded frames. After offloading the computation with the help of the server, the inference time for each frame is significantly reduced. Compared to the previous framework, vehicles will receive more pose information at the same time.

Despite the significant advancements achieved by deep learning-based relocation methods in terms of performance and scene adaptability, certain inherent challenges remain to be addressed. For instance, while the problem of memory resource allocation is mitigated by representing the map as a deep neural network, the reasoning process still relies heavily on computational resources. Considering that relocation modules are frequently deployed on small devices with a high sensitivity to power consumption, such as driverless vehicles, unmanned aerial vehicles, and micro-robots, the inclusion of additional computing hardware would lead to higher energy consumption, thereby hindering the efficient functioning of robots.

In this paper, we present a novel DNN-based relocation framework, focusing on the use of MapNet[22]-based scene building models as exemplary applications. We observed that the reasoning process of MapNet models often involves substantial computing resources on terminal devices, which limits their practical applicability. To overcome this challenge, we propose an offloading approach to improve the efficiency of reasoning while maintaining the accuracy of relocation.

4## 2 Related Work

### 2.1 MapNet Series Camera Relocation Scheme

This series comprises an enhanced version of PoseNet[13] and two newly proposed architectures, MapNet[22] and MapNet++. The backbone network of PoseNet adopts GoogleNet[14], and the representation of the direction when returning to posture is a four-dimensional unit quaternion. The authors of the MapNet series identified shortcomings in this approach, as discussed in their paper. To address these issues, they replaced the backbone network GoogleNet with ResNet34 [23], and changed the direction representation to the logarithm of a unit quaternion.

MapNet is a novel pose regression network that differs from previous pose regression networks that only minimize the absolute pose loss of each image frame. In this method, both the absolute pose loss of each input image frame and the relative pose loss between two image pairs related to this frame are minimized. The learned pose feature distribution using this method has a strong correlation with the actual value, and the combination of absolute and relative loss ensures global consistency in pose estimation, leading to a significant improvement in relocation performance.

Due to the labor-intensive and costly nature of labeling data, a significant amount of unannotated data exists in the real world, such as robot trajectory and GPS data obtained via visual odometer calculations. MapNet++ leverages these unmarked data to fine-tune the supervised training network in a self-supervised manner. Specifically, the network minimizes the sum of the absolute pose loss estimated by the network before and the relative pose loss between frames provided by unmarked data. The efficacy of this approach stems from the high robustness of DNN-based pose regression networks to the scene during the pose estimation process, albeit at the cost of reduced accuracy. In contrast, VO algorithms have high accuracy in estimating the relative pose between frames, but limited generalization capabilities to novel scenes and are prone to bias in the pose estimation process. Similar results can also be inferred from other sensors, such as IMUs. Therefore, fine-tuning the supervised training network with unmarked data leads to better accuracy of pose estimation, which has been demonstrated by the experiments in the original paper.

Upon obtaining the absolute pose of the targeted frame through MapNet++, the proposed approach utilizes the pose graph optimization (PGO)[24] algorithm to fuse the absolute pose with the relative pose between frames obtained by the VO algorithm, thereby establishing a longer constraint on the corresponding trajectory for the targeted frame. This integration of the MapNet++ and PGO algorithms leads to an improvement in the accuracy of pose estimation.

### 2.2 Related Work on Offloading Computing Resource

Several studies have investigated offloading the computing tasks in SLAM systems. Benavidz et al. [25] deployed a cloud-based Robot Operating System (ROS) instance to offload feature recognition and matching stages in the visual SLAM system. Sossalla et al. [26] proposed offloading ORB-SLAM2 computing to an edge server and evaluated the power consumption and network utilization of the SLAM system.

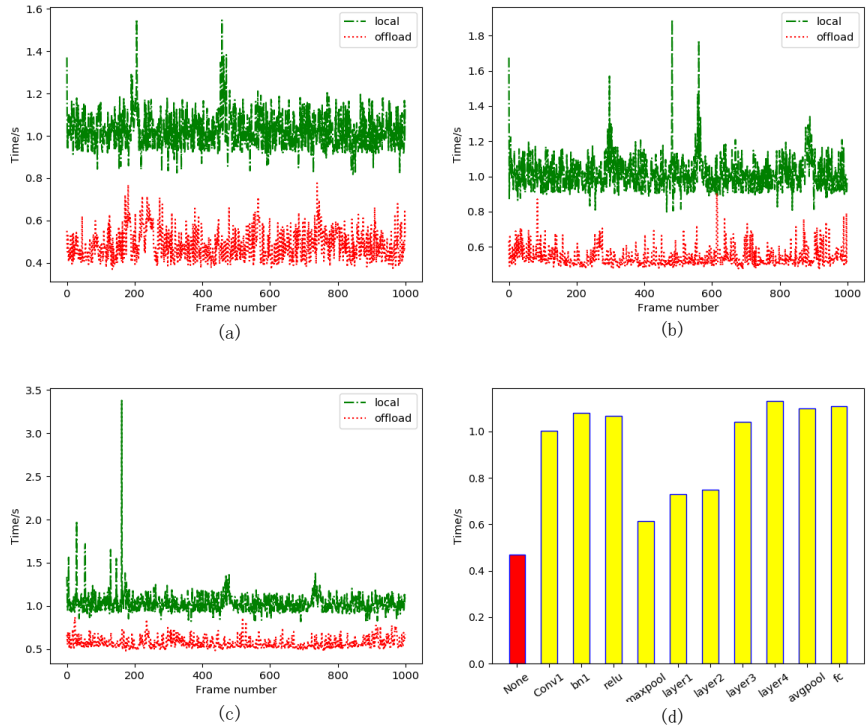

**Fig. 4.** (a): PoseNet (new version) inference time. (b): MapNet inference time. (c): MapNet++ inference time. (d):Network splitting experiments were conducted on the 7Scenes dataset. The abscissa represents the layer at which the network infers termination on mobile devices. For example, relu indicates that the network infers to this layer on the mobile device, and the remaining network inferences are performed at the server. None indicates that only the corresponding frames are reasoned at the server. The ordinate represents the average time spent per frame after reasoning for 100 frames.

Chen et al. [27] developed an edge-assisted vision SLAM system which analyzed the extraction method of key frame subsets, and used the edge server to adaptively select available communication and computing resources to construct the best global map under resource constraints. Sossalla et al. [28] proposed introducing the unloading delay threshold to dynamically control the unloading of the visual SLAM system, which reduces the probability and absolute error of tracking failure and improves CPU utilization. Ali et al. [29] proposed the Edge-SLAM system which uses the edge server to unload the computing resources of the local visual SLAM system. For ORB-SLAM2, local mapping and closed-loop detection are placed on the edge while the tracking part is placed on the mobile device to reduce memory and computational power requirements. Sarker et al. [30] used an intelligent edge gateway based on fog service to handle location and mapping tasks on the SLAM system which are computationally expensive. Wright et al. [31] proposed an output-driven consistency model to divide the SLAM system between the vehicle and edge equipment, maintaining the consistency level while minimizing network traffic and ensuring accurate pose estimation. Xu et al. [32] proposed the SwarmMap framework which can extend the cooperative vision SLAM service in the edge offload setting, greatly reducing data redundancy and positioning errors in the SLAM system. RecSLAM [33], a multi-robot laser SLAM system, uses hierarchical map fusion technology to guide the original data generated by mobile robots to the edge server for real-time fusion, and sends the data to the cloud for fusion. The proposed multi-robot collaborative processing framework optimizes the offloading of robots to the edge and ensures workload balance between edge servers.

The extant methods mainly focus on offloading the computation of the entire SLAM system, encompassing tracking, relocation, closed-loop detection, and mapping. Nevertheless, in several cases, the requirement is not for the complete pipeline but only for specific components that can assist in positioning using available high-precision maps. Furthermore, the existing mainstream offloading strategies are geared towards conventional SLAM systems or robot clusters, and no camera relocation approach based on Deep Neural Networks (DNN) has been proposed for offloading, to the best of our knowledge.

## 3  Methods

In this section, we aim to introduce the DNN-based relocation module and the challenges it encounters, followed by the presentation of the advanced design of the DNN-based relocation module, which constitutes an upgraded version of the MapNet[22] series relocation methods that supports the offloading of the computationally intensive reasoning process to the server. Our goals for this upgraded version are two-fold: first, to enhance the reasoning speed of the network on the autonomous vehicle without requiring high-performance computing equipment; second, to ensure that our proposed method achieves the same accuracy as the original relocation model, particularly in terms of attitude regression. We note that we will not delve into the optimization of vehicle posture using PGO[24] in a sliding window, which is a separate step from the MapNet reasoning model, since it necessitates posture estimation from multiple frames of images. Our focus remains solely on the computation and reasoning involved in the original MapNet series model for autonomous vehicle posture regression.

### 3.1  Overview of DNN-based Relocalization Module

The left part of Fig. 1 depicts the pose reasoning flowchart based on the MapNet[22] series camera relocalization method. It can be observed that the relocalization module of the autonomous vehicle first acquires an environmental image (or frame) via the visual sensor, which is subsequently fed into a deep neural network (DNN)-based scene representation model. The system accepts regular RGB images, though it can also accept color and depth images simultaneously. Ultimately, the network reasoning outputs a pose.

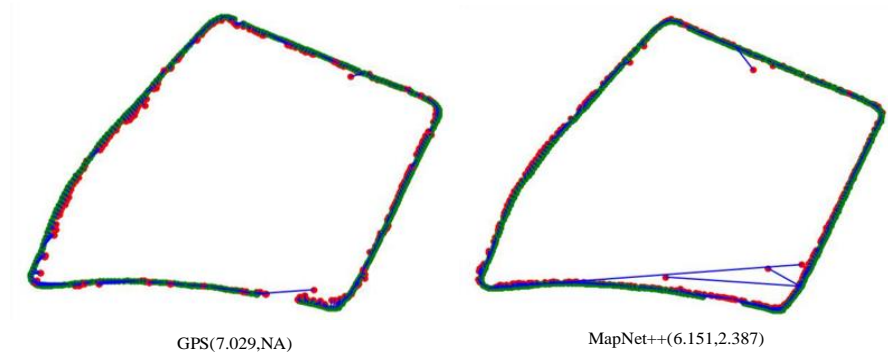

GPS(7.029,NA)　　　　　　　　　　　MapNet++(6.151,2.387)

**Fig. 5.** Comparison of MapNet++ inference results and GPS results of loop sequences in Robot-Car dataset.

### 3.2 Challenges in DNN-based Relocalization Module

The current deep neural network model is composed of multiple hidden layers, and the processing of each frame is carried out sequentially through each layer, resulting in significant computational demands on terminal devices. This is particularly challenging for devices without GPU parallel computing capabilities. During the posture estimation process of autonomous vehicles, the visual sensor captures environmental frames in a sequential manner. If the previous frame has not completed processing within the system, subsequent frames will not be estimated until processing for the previous frame has completed. This results in a delay in obtaining posture data and poses challenges for accurate vehicle posture correction. As shown in Fig. 1, the visual sensor may capture additional image data while processing a particular frame. In practical applications, it is not feasible to wait for the completion of processing for the previous frame before commencing processing for the subsequent frame, as this would not meet real-time processing requirements.

### 3.3 Offloading strategy

We propose a solution to address the time-consuming nature of the relocation module by offloading its computation to the server, while allowing subsequent processing to be carried out on the mobile device. The detailed reasoning process of the upgraded version is depicted in the right part of Fig. 1. We propose a strategy in which the less computationally demanding parts of the module are computed locally, while the more intensive ones are offloaded to the server for processing. The server's higher computing power enables it to complete these calculations efficiently, after which it returns the results to the mobile device. The network layering strategy for the MapNet[22] series of backbone networks is determined by calculating the offload times of several key layers. Our framework is designed to facilitate cloud-edge collaborative computing across three entities: local mobile devices, servers, and LAN. The model is deployed on both local mobile devices and servers. To conduct posture inference based on the offload strategy, we first establish data communication between the mobile device and server.

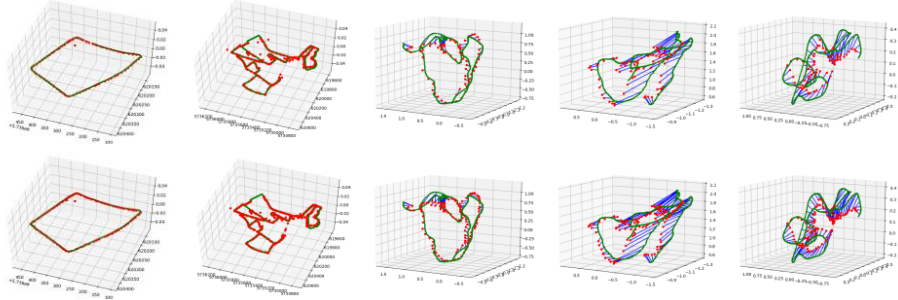

**Fig. 6.** Row 1: Local run frequency results. Row 2: Offloading run frequency results. Green for groundtruth and red for inference result. From left to right are testing sequences:RobotCar-loop,RobotCar-full,7Scenes-Chess-05,7Scenes-Office-06,7Scenes-Heads-01.

Fig. 1 depicts the pipeline of the repositioning module in the upgraded autonomous vehicle proposed in this study. During the reasoning process, the local mobile device's visual sensor initially captures the environmental photograph, which is subsequently preprocessed to conform to the predefined input shape of the network. Meanwhile, the server opens a port to await the unloading request transmitted from the mobile device. Upon receiving the processing request from the mobile terminal, the server executes DNN to reason about the relevant data. Finally, the reasoning result is returned to the mobile device and used for vehicle auxiliary positioning. It is observed that the upgraded relocation pipeline produces more pose data than the approach depicted in Fig. 1 that solely performs reasoning locally, as the reasoning time of a single frame is substantially reduced. While these changes are theoretically straightforward, integrating these aspects to function in harmony presents a significant challenge. While this study proposes a conceptual offload design for the network, integrating the framework into a specific relocation scheme is imperative to determine the appropriate implementation of the offload. To validate the feasibility of our concept, we implemented the relocation framework based on MapNet[22] series.

## 4    Experimental Results

### 4.1    Setup

To evaluate our proposed offloading strategy, we conducted simulation experiments using two different computing devices to represent the mobile and server ends respectively. A low-cost development board, the NVIDIA Jetson Nano development kit with Quad-core ARM A57 CPU, 128-core Maxwell GPU (1.43GHz), and 4GB memory, was used as the computing center for simulating an autonomous vehicle. For the server module, we used a Dell notebook computer equipped with Intel (R) Core (TM) i5-7300HQ CPU (2.50GHz, 4-core), NVIDIA GeForce GTX 1050 GPU, and 8GB memory. The main difference between the server module and the mobile terminal is the higher performance processor of the former. Both devices run on Ubuntu 18.04LTS operating system and are connected to the same LAN. We chose L-

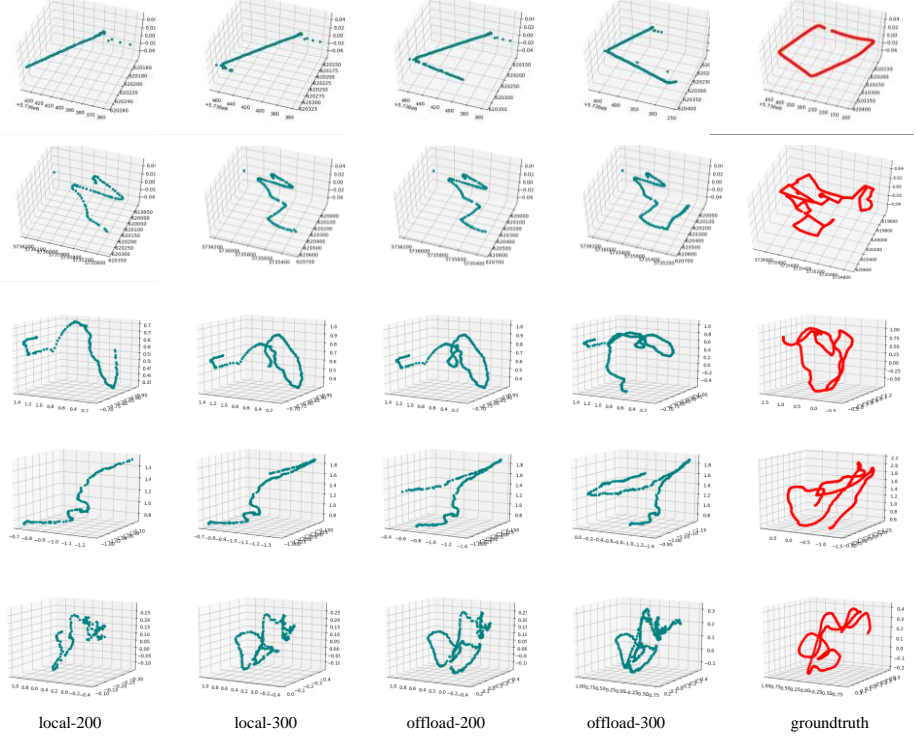

**Fig. 7.** local-200: Run on the mobile device for 200 seconds. local-300: Run on the mobile device for 300 seconds. offload-200: Run using the offloading strategy for 200 seconds. offload-300: Run using the offloading strategy for 300 seconds. From top to bottom are testing sequences:RobotCar-loop,RobotCar-full,7Scenes-Chess-05,7Scenes-Office-06,7Scenes-Heads-01. We provide the groundtruth as a comparison in the fifth column.

AN for data transmission because the focus of this work is on the impact of computing power on system performance.

### 4.2 Dataset

We employed the Oxford Robot dataset [34] and the 7Scenes dataset [35] as the input sources for our experiments, which were also used in the original MapNet[22] series. The Oxford Robot dataset was captured by a vehicle driving twice a week in the center of Oxford, over the course of more than a year. This dataset includes almost 20 million images and encompasses various weather conditions, allowing us to study the performance of autonomous vehicles in real-world and dynamic urban scenes. We selected the loop and full sequences mentioned in the original MapNet series paper, both of which comprise a large number of continuously captured road photos. The 7Scenes dataset includes RGB-D image sequences of seven indoor scenes, which were recorded by handheld Kinect RGB-D cameras. Each sequence in this dataset contains 500-1000 frames, and each frame includes an RGB image, depth image, and pose.

**Table 1.** Single and multi frame network split results on the 7Scenes dataset(null-maxpool)

| local inference to | null | conv1 | bn1 | relu | maxpool |
|---|---|---|---|---|---|
| average of 100 frames time/s | 0.4710 | 1.0022 | 1.0804 | 1.0672 | 0.6140 |
| single frame time/s | 0.5612 | 1.2357 | 1.1516 | 1.5340 | 0.6589 |

**Table 2.** Single and multi frame network split results on the 7Scenes dataset(layer1-end of)

| local inference to | layer1 | layer2 | layer3 | layer4 | avgpool | fc |
|---|---|---|---|---|---|---|
| average of 100 frames time/s | 0.7287 | 0.7480 | 1.0426 | 1.1310 | 1.1010 | 1.1099 |
| single frame time/s | 0.8266 | 0.7595 | 0.8537 | 0.9657 | 0.8700 | 0.8609 |

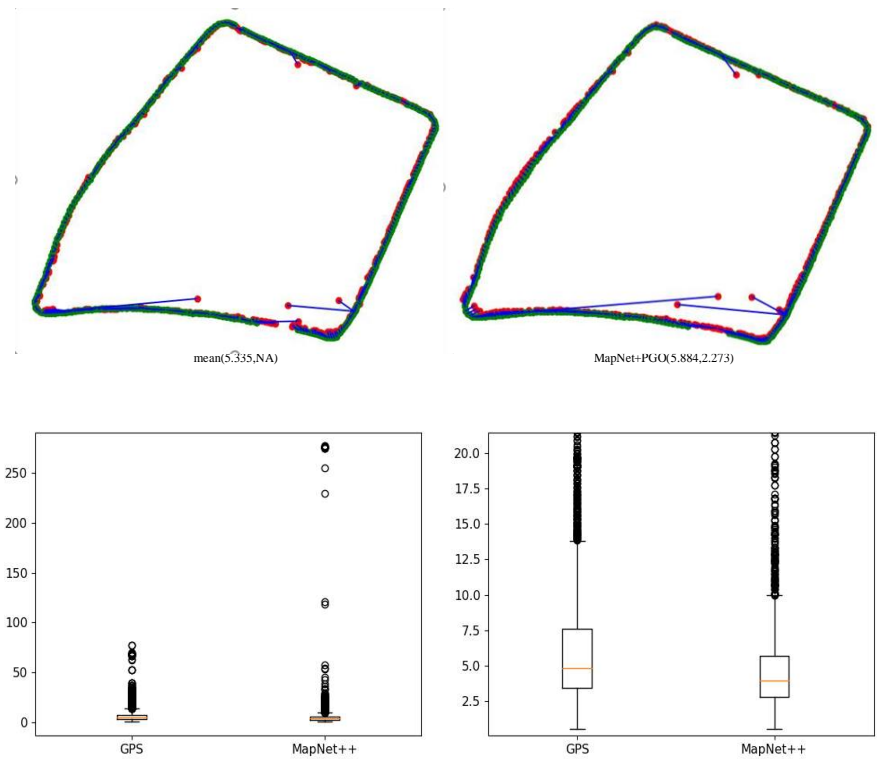

**Fig. 8.** Top: Comparison of the results obtained by averaging MapNet++ and GPS data with MapNet+PGO. Bottom: Loss distribution of GPS data and MapNet++ results.

### 4.3 Experimental details

To ensure the applicability and effectiveness of the proposed framework, we employed a different approach compared to loading data in batches into the network. we utilized OpenCV to read the photos in the dataset frame by frame at the mobile end, which simulates the real-world data reading scenario of autonomous vehicles on the road.

**Network split.** Given that the network structure of MapNet[22] is derived from the PoseNet(new version) network with a ResNet34[23] backbone, and the modification has only a slight impact on the computation, we conducted a network splitting experiment on the middle layer of the PoseNet(new version) network to represent our approach. To account for network transmission rate and fluctuation, we selected 100 consecutive images from the 7Scenes[35] dataset for the experiment and calculated the inference time for each image. The results were averaged to examine the effect of different layers of the neural network on inference time, as show in Fig. 1d). In order to simulate realistic scene, we also conducted single-frame splitting inference. Table 1 and Table 2 summarizes the final results for the two approaches. Our findings demonstrate that unloading all images is more conducive to on-device inference. As the middle layer of the neural network contains a significant number of parameters, unloading data from this layer alone does not produce the desired result. Therefore, we choose to offload the entire image and subsequently upload it to the server for inference.

**Offloading time comparison.** Building upon the current configuration, we select a predetermined number of frames from the dataset for our unloading tests and subsequently compare the outcomes of the two distinct configurations.

*Local pose calculation.* Run PoseNet(new version), MapNet, MapNet++ on Jetson nano.

*Server-side computing.* Use our proposed offloading strategy to unload the three models.

Fig. 1 illustrates the model reasoning time under two different configurations. It is evident that the powerful computing capability of the server has significantly improved the reasoning speed of the model.

### 4.4 Inference in Oxford RobotCar Dataset and 7Scenes Dataset

Based on the aforementioned simulation results, we performed experiments on two datasets, namely RobotCar[34] and 7Scenes[35]. The network model selected for evaluation was the original paper's best-performing model, MapNet++, without any PGO[24] optimization. We evaluated the model's performance from three perspectives: accuracy, route, and inference frequency of two different schemes.

**Accuracy.** As this article aims to address computational efficiency, we have focused our analysis on loop sequences from the RobotCar dataset. As shown in Fig. 1, the results of our reasoning are similar with those reported in the original paper (albeit potentially influenced by variations in machine specifications). Notably, DNN-based scene representation models exhibit smaller translation errors than GPS.

**Inference ferquency.** In the field of autonomous driving, sensor fusion is an advanced approach for assisting vehicle localization. Our upgraded relocation scheme can serve as both a standalone localization module and as a branch of a multi-sensor scheme. The results depicted in Fig. 1 reveal that our enhanced framework has the capability to generate a larger amount of posture data within the same time period. These data can be integrated with the readings from other sensors to correct the attitude of the vehicle, which is of significant practical value.

**Route.** If the model needs to output its pose for each frame, our proposed scheme can achieve a longer trajectory within the same time period. As shown in Fig. 1, a comparison of the reasoning track results between the two schemes demonstrates that our upgraded version is capable of covering a greater distance over time, which further underscores the advantages of our proposed framework.

### 4.5   Discussion on data fusion

In the original paper, PGO[24] was utilized by the author to optimize the results generated by MapNet++. This method led to an average translation error that was even less than the average translation error of GPS. Nevertheless, our observation revealed that after averaging the results of GPS and MapNet++, the final result was better than that of MapNet+PGO. Fig. 1 displays the overall distribution of the error data from both sets. We observed that although MapNet++ had some large outliers, the overall variance was smaller than that of the GPS data. Our analysis revealed that the reason why the average result was better is that the average value can reduce noise, particularly when the error of some frames is too large, leading to more stable overall data. This provides a novel approach for the relocation scheme of autonomous vehicles based on DNN. In practical applications, we can integrate the output results of the network and GPS data to provide additional auxiliary information for the pose correction of autonomous vehicles.

## 5   Conclusion

In this paper, we have presented a novel framework for automatic vehicle relocation based on DNN. Our approach involves offloading the reasoning process to a server, which reduces the computational burden on mobile devices. We have demonstrated the effectiveness of our framework using the MapNet series of relocation schemes. Our experimental results show that our proposed framework can significantly enhance the reasoning efficiency of DNN-based relocation modules in autonomous vehicles. The improved reasoning frequency and route performance highlight the practical significance of our approach. This work also have potential values in some other fields as cloud robotics [36-41]. Future work will focus on extending our offloading strategy to other modules of the SLAM system, particularly the learning-based approach. We are interested in autonomous vehicle positioning and cross-device model inference.

## Acknowledgment

This work was supported by National Natural Science Foundation of China (NSFC) (Grant No. 62162024,62162022),the Key Research and Development Program of Hainan Province (Grant No.ZDYF2021GXJS003，ZDYF2020040),the Major science and technology project of Hainan Province (Grant No. ZDKJ2020012), Hainan Provincial Natural Science Foundation of China (Grant No. 620MS021,621QN211),Science and Technology Development Center of the Ministry of Education Industry-university-Research Innovation Fund (2021JQR017).